\documentclass{Interspeech2024}
\usepackage{multirow}
\usepackage{tablefootnote}
\usepackage[table,xcdraw]{xcolor}
\usepackage{cite}

\interspeechcameraready

\title{Attentive Merging of Hidden Embeddings from Pre-trained Speech Model for Anti-spoofing Detection}

\name[affiliation={1}]{Zihan}{Pan}
\name[affiliation={1}]{Tianchi}{Liu}
\name[affiliation={1}]{Hardik B.}{Sailor}
\name[affiliation={1}]{Qiongqiong}{Wang}

\address{
    $^1$ Institute for Infocomm Research (I$^2$R), Agency for Science, Technology and Research (A$^\star$STAR), Singapore}
\email{{ \{Pan\underline{\enskip}Zihan, liu\underline{\enskip}tianchi, sailor\underline{\enskip}hardik\underline{\enskip}bhupendra, wang\underline{\enskip}qiongqiong\}@i2r.a-star.edu.sg}}

\keywords{anti-spoofing, SSL model, WavLM, attentive merging}

\begin{document}

\maketitle

\begin{abstract}
    

Self-supervised learning (SSL) speech representation models, trained on large speech corpora, have demonstrated effectiveness in extracting hierarchical speech embeddings through multiple transformer layers. However, the behavior of these embeddings in specific tasks remains uncertain. This paper investigates the multi-layer behavior of the WavLM model in anti-spoofing and proposes an attentive merging method to leverage the hierarchical hidden embeddings. Results demonstrate the feasibility of fine-tuning WavLM to achieve the best equal error rate (EER) of 0.65\%, 3.50\%, and 3.19\% on the ASVspoof 2019LA, 2021LA, and 2021DF evaluation sets, respectively. Notably, We find that the early hidden transformer layers of the WavLM large model contribute significantly to anti-spoofing task, enabling computational efficiency by utilizing a partial pre-trained model.
\end{abstract}

\section{Introduction}

Audio spoofing refers to the malicious manipulation of speech or the creation of synthetic speech to deceive listeners or bypass security systems \cite{wang2021prosody}. This poses significant challenges in various real-world applications, such as voice authentication, media integrity, and the detection of fake news \cite{NEURIPS2023_9d276b0a}. For instance, fraudsters may use advanced voice conversion or text-to-speech techniques to impersonate individuals for unauthorized access to confidential information or financial resources \cite{yamagishi2021asvspoof}. Consequently, the development of robust anti-spoofing techniques has become crucial to ensure the security, integrity, and trustworthiness of speech-based systems. Anti-spoofing aims to identify and prevent these fraudulent attempts by distinguishing genuine speech from spoofed audio.

Conventional anti-spoofing methods, while foundational, exhibit limitations in adapting to evolving spoofing tactics. These methods rely on handcrafted acoustic features like MFCCs or LFCCs \cite{sahidullah15_interspeech}, which require expert knowledge and may not capture all necessary discriminative information \cite{chen20_odyssey,xu2021towards}. This leads to generalization challenges, as traditional methods struggle to adapt to novel or unseen spoofing attacks \cite{das2020assessing}. The use of classifiers like Gaussian mixture models or deep Convolutional Neural Netowrks (CNN) \cite{li2021survey} further highlights these limitations, as their adaptability is constrained by the static nature of the handcrafted features they depend on, leaving systems vulnerable to novel spoofing methods \cite{dinkel2017small,alzantot19_interspeech,lai19b_interspeech}. Consequently, there is a need for more robust and flexible anti-spoofing strategies that can autonomously learn and extract relevant features from raw speech data. This transition opens the door for the integration of SSL pre-trained models in anti-spoofing tasks \cite{wang2021antispoofing,tak2021end}, offering a promising approach to overcome the limitations of conventional methods.

SSL models represent a paradigm shift in speech processing, learning rich and nuanced representations from large quantities of unlabeled audio data. Models such as wav2vec 2.0 \cite{baevski2020wav2vec}, HuBERT \cite{hsu2021hubert}, and WavLM \cite{chen2022wavlm} leverage the inherent structure and information within speech signals to develop features that can be fine-tuned for specific downstream tasks, such as speaker verification and speech recognition \cite{fan21_interspeech,yi2021transfer,yang21c_interspeech}. SSL models' ability to capture a broad spectrum of speech characteristics \cite{pasad2021layer} without explicit supervision makes it possible to enhance anti-spoofing measures. Recent research has begun exploring the integration of SSL models within anti-spoofing frameworks, aiming to leverage the sophisticated speech representations these models offer for detecting spoofing attacks. The utilization of wav2vec 2.0's or HuBERT learned features has shown promise in improving the detection accuracy of spoofed speech \cite{wang22_odyssey,martin2022vicomtech,tak22_odyssey}. However, existing works lack a comprehensive exploration of the layer-wise analysis of SSL models in the anti-spoofing task. The study on WavLM \cite{chen2022wavlm} provides an initial exploration into the roles of transformer encoders across various speech processing tasks. However, this exploration leaves several critical questions open for further investigation.

\begin{itemize}
\item How effective are SSL models in detecting spoofing attacks when applied to the anti-spoofing task?
\item From which layer, the speech features extracted by the SSL model are most discriminative in distinguishing between genuine and spoofed speech?
\item Can utilizing a subset of the SSL model's layers yield superior performance compared to the full model?
\end{itemize}

To tackle these questions, we conduct an in-depth analysis of the WavLM large model's architecture and propose an innovative Attentive Merging (AttM) method. This approach is inspired by \cite{huo23b_interspeech} which squeezes the speech features for the ASR task. We combine the embeddings from multiple transformer encoders, which are then fed into a classifier for the anti-spoofing task. Our experimental findings reveal that the proposed method exhibits strong generalization capabilities, even when confronted with unseen and advanced spoofing approaches. Furthermore, our method achieves the SOTA performance while only requiring half the number of SSL encoder layers, resulting in significant computational resource savings.

The remainder of this paper is structured as follows: Section 2 introduces our proposed attentive hidden embedding merging approach for the anti-spoofing task. Section 3 outlines the experimental setup and discusses the evaluation of our proposed methods. In Section 4, we summarize and analyze the experimental results, highlighting the key findings and insights. Finally, Section 5 provides the conclusions.

\section{Methods}

\subsection{WavLM encoder}

WavLM expands upon its predecessors (wav2vec 2.0, HuBERT) by tackling a wider array of speech-processing tasks, including speech separation and speaker identification. It focuses on robustness against background noise, using a large-scale dataset with diverse acoustic environments. WavLM's novel techniques, such as gated relative position bias and data augmentation strategies, significantly improve performance across various speech tasks, demonstrating the model's versatility and the effectiveness of SSL in capturing comprehensive speech representations. So we choose to explore the WavLM's structure to enhance its generalization for anti-spoofing tasks.

WavLM's architecture combines a CNN encoder, functioning as a cochlear filter bank, with transformer \cite{vaswani2017attention} encoders that extract multi-level speech features. The CNN encoder consists of seven temporal convolutional filters, each followed by layer normalization and a GELU activation layer, processing 25ms speech clips with a 20ms stride. We keep the CNN encoder frozen, considering the universality of its extracted acoustic features for anti-spoofing. These features are then processed by 12 or 24 transformer encoders (base or large model). Each transformer block follows HuBERT's design, with 16 attention heads and hidden dimensionality of 768 (base) or 1024 (large).

\begin{figure}[t]
  \centering
  \includegraphics[width=\linewidth]{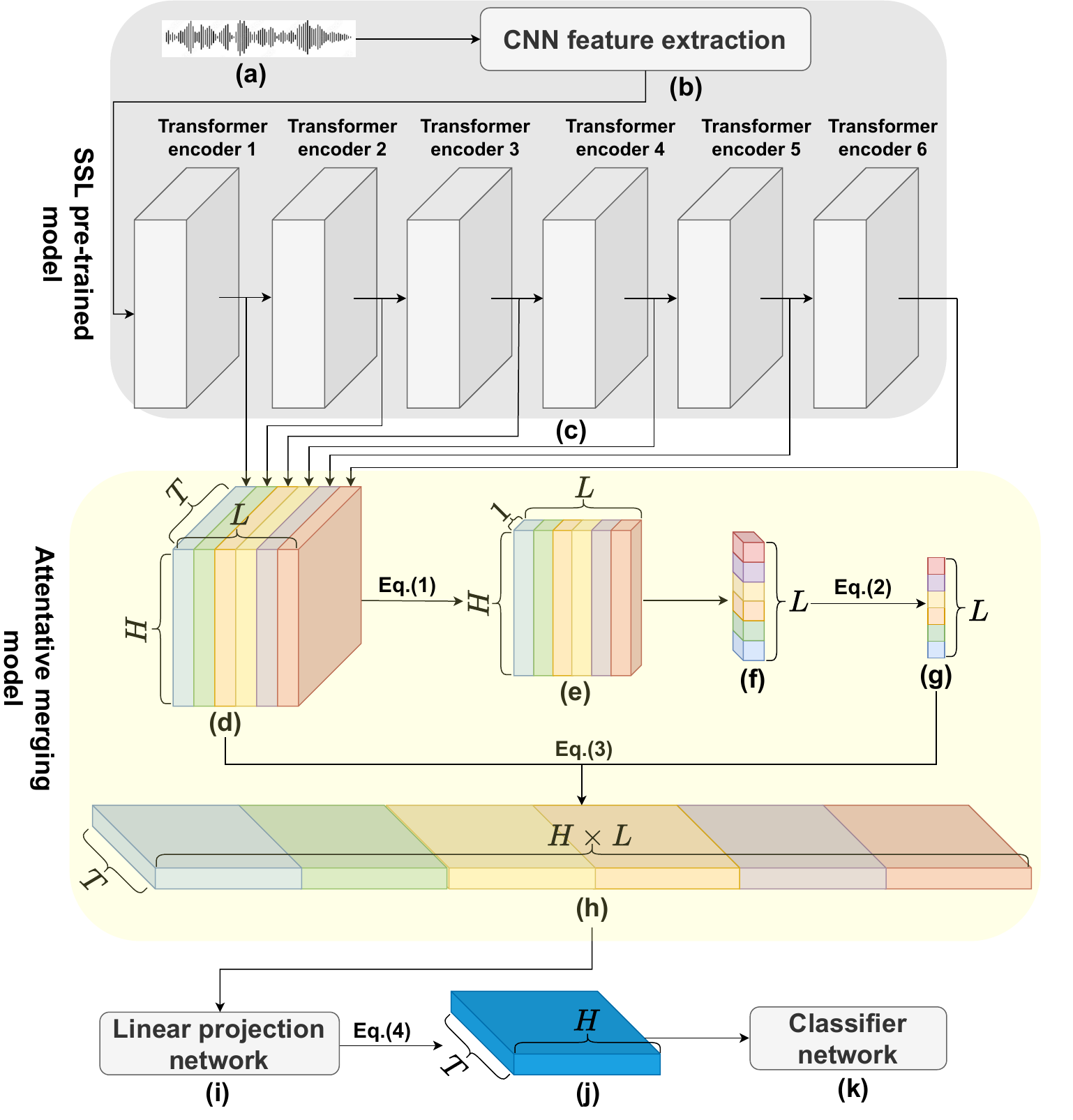}
\caption{Diagram of the attentive merging method for the multi-layer hidden embeddings from WavLM pre-trained model (illustrated for 6 layers). }
\label{fig:method_block}
\end{figure}

\subsection{Proposed AttM of the hidden embeddings}

The hidden embeddings extracted by the SSL pre-trained model capture various levels of speech information, ranging from acoustic features (e.g., phonemes and words) to linguistic aspects (e.g., word meaning and semantics) and speaker characteristics \cite{chen2022wavlm}. Anti-spoofing is a challenging task, as it is not immediately apparent which level of features contributes most significantly to detecting spoofing attacks. To address this, we propose an attentive merging method that emphasizes the most relevant features for anti-spoofing. 

Figure \ref{fig:method_block} illustrates the proposed approach. Given a speech utterance fed into the SSL pre-trained model, a mel-cepstrum-like feature $\mathbf{S} \in \Re^{D \times T}$ is initially obtained from a CNN network (see Figure \ref{fig:method_block}(b)), where $D$ and $T$ represent the number of frequency bins and time frames, respectively. The spectrum is then processed by the tandem transformer encoder, and the hidden embedding tensor from transformer layer $l$ is denoted as $\mathbf{X}_l \in \Re^{T \times H}, l = 1,2,3,...L$, where $L$ is the number of transformer layers and $H$ is the hidden dimension. For instance, $L=24$ and $H=1024$ for the WavLM large model. Subsequently, the stack of all the hidden embeddings $\mathbf{X}^{T\times H\times L}$ (see Figure \ref{fig:method_block}(d)) is squeezed by a two-step process:
\begin{equation}
    \mathbf{x}_{\text{sq}} = \Theta\left[\frac{\Sigma_{t}\mathbf{X}}{T}\mathbf{w}_{\text{sq}}^{H \times 1}\right]
    \label{eq:squeeze}
\end{equation}
First, the information is averaged in $\frac{\Sigma_{t}\mathbf{X}}{T}$ across the time dimension $T$ for each embedding (see Figure \ref{fig:method_block}(e)). Then, the hidden dimension $H$ is squeezed using a fully connected layer with weight tensor $\mathbf{w}_{\text{sq}}$, followed by a SWISH activation function $\Theta$ \cite{ramachandran2017searching}. The resulting $1\times L \times 1$ tensor $\mathbf{x}_{\text{sq}}$ (see Figure \ref{fig:method_block}(f)) encapsulates the spatial-temporal information relevant to the anti-spoofing task. Next, we obtain the attentive weights $\mathbf{x}^{1\times L}_{\text{AttW}}$ (see Figure \ref{fig:method_block}(g)) for each embedding layer by exciting the squeezed tensor:
\begin{equation}
    \mathbf{x}_{\text{AttW}}=\sigma\left[\Theta(\mathbf{x}_{\text{sq}}\mathbf{W}^{L\times s}_{\text{ex1}})\mathbf{W}^{s \times L}_{\text{ex2}}\right]
    \label{eq:attention weight}
\end{equation}
where $\sigma$ denotes the sigmoid activation function. The excitation weights $\mathbf{W}^{L\times s}_{\text{ex1}}$ and $\mathbf{W}^{s \times L}_{\text{ex2}}$ incorporate a scaling vector $s$, which is set to $L/2$ in our experiments.
The attentive weights are applied to the stacked tensor $\mathbf{X}$ along the embedding layer dimension $L$ through a Hadamard product $\circ$, resulting in a re-weighted stack of embedding $\mathbf{X}^{T\times H\times L}_{\text{Att}}$:
\begin{equation}
    \mathbf{X}_{\text{Att}} = \mathbf{X} \circ \mathbf{x}_{\text{AttW}}
    \label{eq:excitation}
\end{equation}
Finally, we concatenate all the embeddings (see Figure \ref{fig:method_block}(h)) and merge them using a 3-layer linear projection network with weights $\mathbf{W}^{(H\times L)\times i}_{\text{L1}}$, $\mathbf{W}^{i\times i}_{\text{L2}}$, and $\mathbf{W}^{i\times H}_{\text{L3}}$. The projection weights serve as a bottleneck network, in which the intermediate dimension $i$ is set as $(H\times L)/4$:
\begin{equation}
    \mathbf{X}_{\text{AttM}} = \mathbf{X}_{\text{Att}}\mathbf{W}_{\text{L1}}\mathbf{W}_{\text{L2}}\mathbf{W}_{\text{L3}}
    \label{eq: attentive merging}
\end{equation}
The resulting attentively merged tensor $\mathbf{X}^{T \times H}_{\text{AttM}}$ (see Figure \ref{fig:method_block}(j)) retains the global information from the spatial-temporal stacked tensor while emphasizing the most relevant transformer hidden embeddings for anti-spoofing. This feature representation will be utilized in the downstream anti-spoofing tasks.


\section{Experiment Setup}

\subsection{Training and evaluation dataset}

For our experiments, we leveraged the diverse and challenging ASVspoof datasets. Specifically, we used the ASVspoof 2019 \cite{nautsch2021asvspoof} Logical Access (19LA) training set for model training and the 19LA development set for validation. To rigorously evaluate the performance and generalization capabilities of our proposed anti-spoofing system, we conducted evaluations on three distinct sets:
\begin{itemize}
    \item The ASVspoof 2019 LA evaluation set \cite{nautsch2021asvspoof} , comprises bona fide and spoofed utterances generated by various text-to-speech (TTS) and voice conversion (VC) algorithms.
    \item The ASVspoof 2021  Logical Access (21LA) evaluation set \cite{yamagishi2021asvspoof}, features a diverse range of state-of-the-art TTS and VC attacks generated using the latest deep learning techniques, designed to test the limits of spoofing countermeasures in a realistic scenario.
    \item The ASVspoof 2021 Deepfake (21DF) evaluation set \cite{yamagishi2021asvspoof}, introduces a more advanced class of spoofing attacks generated by deepfake synthesis techniques, representing the evolving nature of spoofing threats.
\end{itemize}

By evaluating our anti-spoofing system on these diverse evaluation sets, we aim to thoroughly assess its performance against a wide range of spoofing attacks, including those generated by traditional TTS and VC methods, as well as the more recent and sophisticated deepfake synthesis techniques.

\begin{table}[t]
\centering
\caption{Experimental setups. Two training strategies are applied: fix the pre-trained model or fine-tune it.}
\scriptsize
\footnotesize
\setlength{\tabcolsep}{0.1mm}
\renewcommand{\arraystretch}{1.0} 
\begin{tabular}{c|c|c|c}
\hline
\toprule
\quad Pre-trained model \quad & \quad Merging model \quad & \quad Classifier \quad & \quad Strategy \quad\\
\hline
\multirow{3}{*}{WavLM large} & LinM & \multirow{2}{*}{LSTM}  & \multirow{2}{*}{fine-tuned}\\
                       & AttM &                      \\
                       & -          & ECAPA     &   fixed      \\
\hline
\bottomrule
\end{tabular}
\label{tab:model table}
\end{table}

\subsection{Models and tasks}

 Our experiments comprise three primary components: the transformer encoders (Figure \ref{fig:method_block}(c)), the embedding merging block (Figure \ref{fig:method_block}(d-h)), and the classifier (Figure \ref{fig:method_block}(j)). We investigated various combinations of these components using different models, as outlined in Table \ref{tab:model table}. AttM denotes the Attentive Merging approach introduced in Section 2.2, while LinM represents a trainable Linear layer with a positive weight vector $W_{Lin}^{1 \times L} = w_1, w_2, \dots, w_L, L=24$. The LinM layer is designed to attentively weigh and linearly combine the 24 hidden embeddings for the downstream task, similar to the approach employed in \cite{chen2022wavlm}:
\begin{equation}
    X_{\text{LinM}} = \Sigma_l^L(w_l X_l)
    \label{eq:linear weight}
\end{equation}
where $X_l$ represents the hidden embedding from the $l$-th transformer layer, and $X_{\text{LinM}}$ denotes the merged output. The purpose of this merging approach is to examine the contribution of different hidden embeddings to the downstream task.

The merged embedding, $X_{\text{AttM}}$ or $X_{\text{LinM}}$, serves as input to a classifier, which consists of either a single layer of LSTM or an ECAPA-TDNN model \cite{desplanques20_interspeech}. We omit the filter bank blocks from the ECAPA-TDNN model since the pre-trained model has already performed feature extraction. Instead, we employ the ECAPA-TDNN model to classify the output embedding.




\subsection{Fine-tuning the big model with warm-up strategy}


We apply a warm-up scheduler for fine-tuning the WavLM model. There are 3 stages in the training:

\begin{itemize}
    \item [1)] Warm up the learning rate: the learning rate linearly increases in the beginning 5 epochs. To accelerate the training stage, we freeze the WavLM model (Figure\ref{fig:method_block}(b)(c)) but update the attention blocks (Figure\ref{fig:method_block}(d) to (h)) and the classifier (Figure\ref{fig:method_block}(j) from scratch. 
    \item [2)] Learning rate decaying: starting from the $\text{6}^{\text{th}}$ epoch, the learning rate decays exponentially in the following epochs. 
    \item [3)] Start tuning the SSL model: the transformer encoder blocks (Figure\ref{fig:method_block}(c)) are updated from the $\text{11}^{\text{th}}$ epochs.
\end{itemize}

\section{Results and discussion}

\begin{table*}[t] 
\centering
\caption{Experiment results EER(\%) with the fixted WavLM Large model. The SSL layer number denotes how many transformer encoders from the first are utilized. For the baseline models (last two rows) in which there are no hidden embedding merging blocks, the embedding from that particular encoder layer is directly fed into the classifier (-LSTM or -ECAPA). The lowest averaged EER across each model is highlighted in a grey-color box.}
\vspace{-0.1 in}

\footnotesize
\setlength{\tabcolsep}{0.4mm}{
\renewcommand{\arraystretch}{1.0}{
\begin{tabular}{lcccccccccccccccccccc} 
\hline
\toprule
\multicolumn{1}{l}{SSL layer} & \multicolumn{4}{c}{\underline{\qquad\qquad 24 \qquad\qquad}} & \multicolumn{4}{c}{\underline{\qquad\qquad 18 \qquad\qquad}} & \multicolumn{4}{c}{\underline{\qquad\qquad 12 \qquad\qquad}} & \multicolumn{4}{c}{\underline{\qquad\qquad 10 \qquad\qquad}} & \multicolumn{4}{c}{\underline{\qquad\qquad\quad 6 \qquad\qquad}} \\
Dataset & 19LA & 21LA & 21DF & Avg. & 19LA & 21LA & 21DF & Avg. & 19LA & 21LA & 21DF & Avg. & 19LA & 21LA & 21DF & Avg. & 19LA & 21LA & 21DF & Avg. \\
\hline
LinM-LSTM & 1.12 & 11.35 & 5.95 & 6.55 & 0.54 & 7.94 & 4.88 & 5.07 & 0.43 & 7.55 & 4.75 & \cellcolor{gray!25}4.89 & 0.37 & 8.22 & 5.42 & 5.49 & 1.03 & 13.60 & 10.73 & 10.37 \\

LinM-ECAPA & 0.15 & 6.38 & 6.79 & 6.08 & 0.45 & 7.18 & 6.33 & 5.94 & 0.20 & 3.26 & 6.60 & \cellcolor{gray!25}5.33 & 0.24 & 4.79 & 6.52 & 5.58 & 0.63 & 9.30 & 10.33 & 9.21 \\

AttM-LSTM & 0.80 & 6.29 & 6.26 & 5.74 & 0.32 & 5.80 & 5.46 & \cellcolor{gray!25}5.04 & 0.37 & 7.75 & 6.29 & 6.01 & 0.67 & 13.79 & 7.98 & 8.43 & 1.19 & 17.70 & 15.81 & 14.79 \\

AttM-ECAPA & 0.57 & 7.03 & 6.91 & 6.33 & 0.46 & 5.62 & 7.59 & 6.53 & 0.43 & 5.60 & 7.54 & 6.48 & 1.49 & 8.27 & 5.80 & \cellcolor{gray!25}5.88 & 1.41 & 13.86 & 16.02 & 14.21 \\

-LSTM & 9.02 & 16.29 & 14.08 & 14.03 & 5.95 & 13.45 & 11.99 & 11.70 & 1.71 & 8.65 & 6.92 & 6.76 & 1.05 & 10.20 & 5.94 & \cellcolor{gray!25}6.31 & 0.66 & 8.97 & 7.22 & 6.94 \\

-ECAPA & 2.99 & 12.72 & 9.87 & 9.78 & 1.30 & 9.34 & 9.54 & 8.72 & 0.96 & 8.97 & 7.50 & 7.17 & 0.31 & 4.53 & 6.65 & \cellcolor{gray!25}5.63 & 0.69 & 8.05 & 5.69 & 5.68 \\
\bottomrule
\hline
\end{tabular}
}}
\label{tab:fixed_results}
\end{table*}


\begin{table*}[t]
\footnotesize
\centering
\caption{Experiment results EER(\%) with fine-tuned WavLM Large model. Same setting are applied as Table \ref{tab:fixed_results}}
\vspace{-0.1 in}

\setlength{\tabcolsep}{0.4mm}{
\renewcommand{\arraystretch}{1.0}{
\begin{tabular}{lcccccccccccccccccccc}
\hline
\toprule
\multicolumn{1}{l}{SSL layer} & \multicolumn{4}{c}{\underline{\qquad\qquad 24 \qquad\qquad}} & \multicolumn{4}{c}{\underline{\qquad\qquad 18 \qquad\qquad}} & \multicolumn{4}{c}{\underline{\qquad\qquad 12 \qquad\qquad}} & \multicolumn{4}{c}{\underline{\qquad\qquad 10 \qquad\qquad}} & \multicolumn{4}{c}{\underline{\qquad\qquad\quad 6 \qquad\qquad}} \\

Dataset & 19LA & 21LA & 21DF & Avg. & 19LA & 21LA & 21DF & Avg. & 19LA & 21LA & 21DF & Avg. & 19LA & 21LA & 21DF & Avg. & 19LA & 21LA & 21DF & Avg. \\
\hline
LinM-LSTM & 0.37 & 4.21 & 4.92 & 4.35 & 0.62 & 6.08 & 5.18 & 4.92 & 0.54 & 5.84 & 3.46 & \cellcolor{gray!30}3.65 & 0.54 & 4.52 & 4.37 & 4.03 & 0.66 & 14.80 & 10.74 & 10.58 \\
LinM-ECAPA & 0.50 & 6.03 & 6.04 & 5.51 & 0.67 & 7.63 & 7.18 & 6.65 & 0.39 & 4.37 & 7.43 & 6.16 & 0.69 & 4.75 & 5.81 & \cellcolor{gray!30}5.11 & 0.83 & 8.53 & 10.10 & 8.91 \\
AttM-LSTM & 0.85 & 4.46 & 4.73 & 4.31 & 1.46 & 2.84 & 4.10 & 3.60 & 0.65 & 3.50 & 3.19 & \cellcolor{gray!30}3.01 & 1.98 & 5.79 & 6.09 & 5.64 & 1.26 & 12.41 & 10.53 & 10.02 \\
AttM-ECAPA & 0.86 & 5.52 & 5.27 & \cellcolor{gray!30}5.06 & 0.63 & 8.24 & 4.88 & 5.13 & 0.44 & 4.45 & 6.52 & 5.53 & 0.57 & 11.65 & 4.75 & 5.71 & 1.54 & 13.08 & 11.52 & 10.88 \\
-LSTM & 0.30 & 21.67 & 5.35 & 8.08 & 0.50 & 16.84 & 6.03 & 7.63 & 0.98 & 6.86 & 3.30 & \cellcolor{gray!30}3.78 & 1.94 & 6.54 & 4.17 & 4.42 & 1.12 & 7.14 & 6.40 & 6.04 \\
-ECAPA & 0.34 & 7.24 & 5.59 & 5.41 & 0.28 & 7.00 & 7.61 & 6.79 & 0.40 & 4.33 & 7.64 & 6.30 & 0.35 & 3.24 & 4.58 & \cellcolor{gray!30}4.62 & 0.73 & 7.27 & 6.78 & 6.30 \\
\bottomrule
\hline
\end{tabular}
}}
\label{tab:finetune_results}
\vspace{-0.1 in}
\end{table*}

\begin{table}[t]
\centering
\caption{Performance from other works for ASVspoof datasets.}
\footnotesize
\setlength{\tabcolsep}{0.6mm}{
\renewcommand{\arraystretch}{1.0}{
\begin{tabular}{llrrr}
\hline
\toprule
\multicolumn{2}{c}{Model name} & \multicolumn{3}{c}{EER (\%)} \\
\cline{3-5}
\multicolumn{2}{c}{} & 19LA & 21LA & 21DF \\
\hline
\midrule

\multirow{10}{*}{\rotatebox{90}{Public}} & W2V-XLSR-LLGF\cite{wang22_odyssey} & 2.80 & 7.26 & 6.68 \\
                         & HuBERT-XL\cite{wang22_odyssey} & 3.55 & 9.55 & 13.07 \\
                         & W2V-Large1-LLGF\cite{wang22_odyssey} & 0.86 & 13.19 & 7.44 \\
                         & W2V2-LCNN-BLSTM\cite{wang22_odyssey} & - & 7.18 & 5.44 \\
                         & W2V2-base-DARTS\cite{wang2022fully} & 1.19 & 8.16 & - \\
                         & W2V2-large-DARTS\cite{wang2022fully} & 1.08 & 7.86 & - \\
                         & W2V-XLS-128\cite{martin2022vicomtech} & - & 3.54 & 6.18 \\
                         & W2V-XLS-53\cite{martin2022vicomtech} & - & 4.98 & 6.99 \\
                         & ResNet(Ensemble)\cite{chen21b_asvspoof} & - & 3.21 & 16.05 \\
                         & ECAPA-TDNN(Ensemble)\cite{chen21_asvspoof} & - & 5.46 & 20.33 \\
                    
\hline

\multirow{2}{*}{\rotatebox{90}{Ours}} & 12L-WavLM-Large AttM-LSTM & 0.65 & 3.50 & 3.19 \\
                      & 10L-WavLM-Large LinM-LSTM & 0.54 & 4.52 & 4.37 \\

\hline

\bottomrule
\hline
\end{tabular}
}}
\label{tab:compare_results}
\end{table}


\subsection{Weighted hidden embeddings}

\begin{figure}[t]
  \centering
  \includegraphics[width=\linewidth]{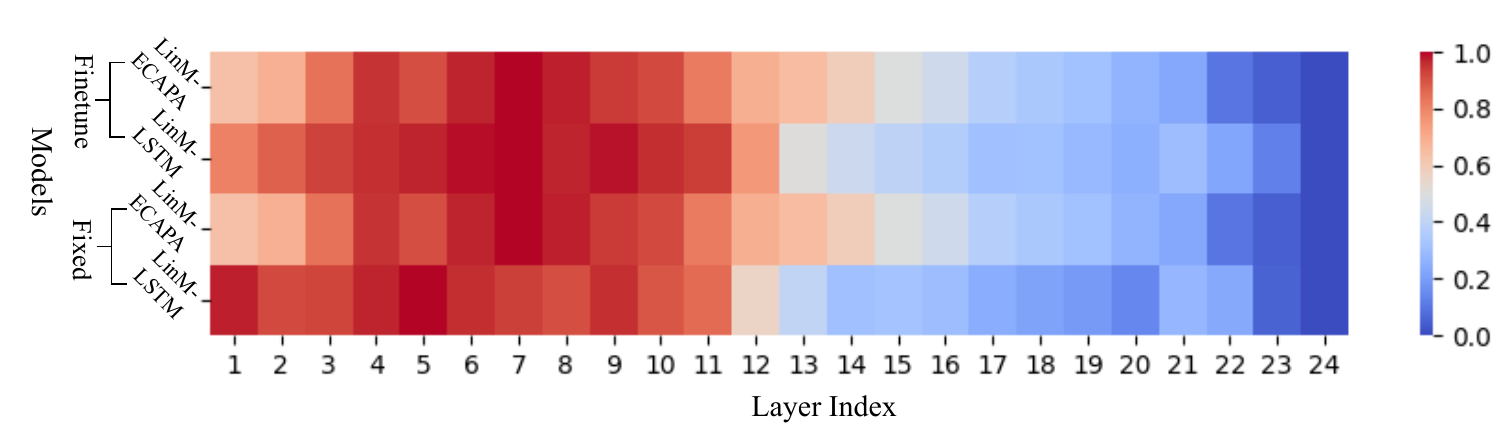}
\caption{Normalized linear weights on the hidden embeddings from the 24 transformer encoders in the WavLM large model.}
\label{fig:weight plot}
\end{figure}



To investigate the contribution of different transformer layers to the anti-spoofing task, we trained two models: LinM-LSTM and LinM-ECAPA, both utilizing the WavLM large model as the front-end. Additionally, we also fine-tuned pre-trained model as described in Section 3.3.

 Figure \ref{fig:weight plot} illustrates the distribution of normalized weights $W_{\text{Lin}}(w_l)$ in (\ref{eq:linear weight}). Our analysis reveals that the trained linear weights $W_{\text{Lin}}$ exhibit similar distributions across various classifiers and training strategies fine-tuned or fixed). Notably, we observe that the hidden embeddings from the early transformer layers up to the middle (around layer 12) make a more significant contribution to the anti-spoofing task compared to the outputs from all layers combined. This finding suggests that the early transformer layers of the WavLM model capture important features and information that are particularly relevant for detecting spoofing attacks.

\subsection{Train and fine-tune strategies for WavLM}

To validate the findings from the previous section, we evaluated additional systems where the WavLM model was partially employed, using a subset of transformer layers ranging from the initial layer to the 6th, 10th, 12th, 18th, and 24th layers. The embeddings from these layers were merged, while the upper embeddings were discarded. Initially, we fixed the WavLM large model and trained only the downstream embedding merging block and classifier, resulting in the EER\% presented in Table \ref{tab:fixed_results}. Then we train and fine-tune the entire system (see the results in Table \ref{tab:finetune_results}). The average error rates (Avg.) across the 3 evaluation datasets were calculated to facilitate comparisons between different numbers of transformer layers.

We observe that fine-tuned WavLM models consistently outperform the fixed WavLM models across various SSL layer configurations and datasets, indicating that fine-tuning adapts the pre-trained representations to the specific anti-spoofing task. In most cases, utilizing an intermediate number of layers (e.g., 12 or 10) yields better results compared to using all 24 layers or only 6 layers, suggesting that intermediate layers capture more relevant information for anti-spoofing. The choice of classifier (LSTM or ECAPA) impacts performance, with LSTM generally outperforming ECAPA, particularly when combined with LinM or AttM embedding merging models. The AttM-based model often surpasses the LinM-based model, indicating that AttM's attention mechanism effectively prioritizes informative embeddings for anti-spoofing. Performance varies across datasets (19LA, 21LA, 21DF), with 21LA and 21DF being more challenging, as evidenced by higher EERs. 


These results highlight the importance of considering the layer-wise contributions when leveraging pre-trained models for anti-spoofing tasks. By focusing on the intermediate layers, which appear to encode more discriminative information, we can potentially improve the performance and efficiency of anti-spoofing systems. This insight can guide future research and development efforts in optimizing the utilization of pre-trained models for anti-spoofing applications.

\subsection{Compared with literature}

Table \ref{tab:compare_results} presents a comparison of EERs for various models on the ASVspoof datasets. The models are divided into two categories: public models from the literature and the proposed models (referred to as ``Ours"). Among the public models, the W2V-Large1-LLGF model achieves the lowest EER of 0.86\% on the 19LA subset, while the W2V-XLS-128 model obtains the best EER of 3.54\% on the 21LA subset. For the 2021DF subset, the W2V2-LCNN-BLSTM model demonstrates the lowest EER of 5.44\% among the public models. In comparison, the proposed models, namely the 12L-WavLM-Large AttM-LSTM and 10L-WavLM-Large LinM-LSTM, show competitive performance across all three subsets. The 12L-WavLM-Large AttM-LSTM model achieves EERs of 0.65\%, 3.50\%, and 3.19\% on the 19LA, 21LA, and 21DF subsets, respectively. The 10L-WavLM-Large LinM-LSTM model obtains EERs of 0.54\%, 4.52\%, and 4.37\% on the same subsets. Moreover, we achieve the SOTA EER for the most challenging DF evaluation set in the ASVspoof 2021 dataset.

The comparison highlights that the proposed models, which utilize WavLM-Large with specific layer configurations and embedding merging techniques (AttM-LSTM and LinM-LSTM), outperform the state-of-the-art public models on the ASVspoof datasets. Moreover, we use a partial of the pre-trained model to save the computational resources, other than the entire parameter-heavy model.


\section{Conclusion}

We propose an attentive hidden embedding merging technique and address the questions raised in this study. Our experiments show that the WavLM model exhibits anti-spoofing capabilities, even without fine-tuning, which can be harnessed by a simple downstream classifier. The early hidden layers of the WavLM model contribute the most to the performance, suggesting that the acoustic, lexical, and sentiment features captured by these layers are highly discriminative for detecting spoofing attacks. By strategically merging the hidden embeddings from these informative layers, we achieved superior performance compared to using the entire pre-trained model, outperforming other state-of-the-art systems on the ASVspoof datasets. Moreover, our approach offers computational advantages by requiring only a subset of the parameter-heavy pre-trained model. In conclusion, our study highlights the effectiveness of pre-trained speech representation models for anti-spoofing tasks and the importance of considering layer-wise contributions. The proposed attentive merging technique presents an efficient approach to leverage pre-trained models for enhanced anti-spoofing performance.

\section{Acknowledgements}
{\fontsize{9}{9}\selectfont
This work is supported
by the National Research Foundation, Prime Minister’s Office, Singapore, and the Ministry of Communications and Information, under its Online Trust and Safety (OTS) Research Programme (MCI-OTS-001). Any opinions, findings and conclusions or recommendations expressed in this material are those of the author(s) and do not reflect the views of National Research Foundation, Prime Minister’s Office,  Singapore, or the Ministry of Communications and Information.}

\bibliographystyle{IEEEtran}
\bibliography{mybib}

\begin{thebibliography}{10}
\providecommand{\url}[1]{#1}
\csname url@samestyle\endcsname
\providecommand{\newblock}{\relax}
\providecommand{\bibinfo}[2]{#2}
\providecommand{\BIBentrySTDinterwordspacing}{\spaceskip=0pt\relax}
\providecommand{\BIBentryALTinterwordstretchfactor}{4}
\providecommand{\BIBentryALTinterwordspacing}{\spaceskip=\fontdimen2\font plus
\BIBentryALTinterwordstretchfactor\fontdimen3\font minus \fontdimen4\font\relax}
\providecommand{\BIBforeignlanguage}[2]{{%
\expandafter\ifx\csname l@#1\endcsname\relax
\typeout{** WARNING: IEEEtran.bst: No hyphenation pattern has been}%
\typeout{** loaded for the language `#1'. Using the pattern for}%
\typeout{** the default language instead.}%
\else
\language=\csname l@#1\endcsname
\fi
#2}}
\providecommand{\BIBdecl}{\relax}
\BIBdecl

\bibitem{wang2021prosody}
T.~Wang, R.~Fu, J.~Yi, J.~Tao, Z.~Wen, C.~Qiang, and S.~Wang, ``Prosody and voice factorization for few-shot speaker adaptation in the challenge m2voc 2021,'' in \emph{ICASSP 2021-2021 IEEE International Conference on Acoustics, Speech and Signal Processing (ICASSP)}.\hskip 1em plus 0.5em minus 0.4em\relax IEEE, 2021, pp. 8603--8607.

\bibitem{NEURIPS2023_9d276b0a}
T.~Liu, K.~A. Lee, Q.~Wang, and H.~Li, ``Disentangling voice and content with self-supervision for speaker recognition,'' in \emph{Advances in Neural Information Processing Systems}, vol.~36, 2023, pp. 50\,221--50\,236.

\bibitem{yamagishi2021asvspoof}
J.~Yamagishi, X.~Wang, M.~Todisco, M.~Sahidullah, J.~Patino, A.~Nautsch, X.~Liu, K.~A. Lee, T.~Kinnunen, N.~Evans \emph{et~al.}, ``Asvspoof 2021: accelerating progress in spoofed and deepfake speech detection,'' in \emph{ASVspoof 2021 Workshop-Automatic Speaker Verification and Spoofing Coutermeasures Challenge}, 2021.

\bibitem{sahidullah15_interspeech}
M.~Sahidullah, T.~Kinnunen, and C.~Hanilçi, ``{A comparison of features for synthetic speech detection},'' in \emph{Proc. Interspeech 2015}, 2015, pp. 2087--2091.

\bibitem{chen20_odyssey}
T.~Chen, A.~Kumar, P.~Nagarsheth, G.~Sivaraman, and E.~Khoury, ``{Generalization of Audio Deepfake Detection},'' in \emph{Proc. The Speaker and Language Recognition Workshop (Odyssey 2020)}, 2020, pp. 132--137.

\bibitem{xu2021towards}
C.~Xu, M.~Zhang, J.~Li, N.~Poh, and G.~Zhao, ``Towards generalization of spoofing countermeasures: Review and insights,'' \emph{IEEE Access}, vol.~9, pp. 152\,693--152\,718, 2021.

\bibitem{das2020assessing}
R.~K. Das, J.~Yang, and H.~Li, ``Assessing the scope of generalized countermeasures for anti-spoofing,'' in \emph{ICASSP 2020-2020 IEEE International Conference on Acoustics, Speech and Signal Processing (ICASSP)}.\hskip 1em plus 0.5em minus 0.4em\relax IEEE, 2020, pp. 6589--6593.

\bibitem{li2021survey}
Z.~Li, F.~Liu, W.~Yang, S.~Peng, and J.~Zhou, ``A survey of convolutional neural networks: analysis, applications, and prospects,'' \emph{IEEE transactions on neural networks and learning systems}, vol.~33, no.~12, pp. 6999--7019, 2021.

\bibitem{dinkel2017small}
H.~Dinkel, Y.~Qian, and K.~Yu, ``Small-footprint convolutional neural network for spoofing detection,'' in \emph{2017 International Joint Conference on Neural Networks (IJCNN)}.\hskip 1em plus 0.5em minus 0.4em\relax IEEE, 2017, pp. 3086--3091.

\bibitem{alzantot19_interspeech}
M.~Alzantot, Z.~Wang, and M.~B. Srivastava, ``{Deep Residual Neural Networks for Audio Spoofing Detection},'' in \emph{Proc. Interspeech 2019}, 2019, pp. 1078--1082.

\bibitem{lai19b_interspeech}
C.-I. Lai, N.~Chen, J.~Villalba, and N.~Dehak, ``{ASSERT: Anti-Spoofing with Squeeze-Excitation and Residual Networks},'' in \emph{Proc. Interspeech 2019}, 2019, pp. 1013--1017.

\bibitem{wang2021antispoofing}
X.~Wang, C.~Wu, S.~Wang, L.~Xie, J.~Pu, Q.~Peng, and T.~Kinnunen, ``Anti-spoofing in speaker verification: An overview of modern approaches,'' in \emph{IEEE Spoken Language Technology Workshop (SLT)}.\hskip 1em plus 0.5em minus 0.4em\relax IEEE, 2021, pp. 1316--1323.

\bibitem{tak2021end}
H.~Tak, J.~Patino, M.~Todisco, A.~Nautsch, N.~Evans, and A.~Larcher, ``End-to-end anti-spoofing with rawnet2,'' in \emph{ICASSP 2021-2021 IEEE International Conference on Acoustics, Speech and Signal Processing (ICASSP)}.\hskip 1em plus 0.5em minus 0.4em\relax IEEE, 2021, pp. 6369--6373.

\bibitem{baevski2020wav2vec}
A.~Baevski, Y.~Zhou, A.~Mohamed, and M.~Auli, ``wav2vec 2.0: A framework for self-supervised learning of speech representations,'' \emph{Advances in Neural Information Processing Systems}, vol.~33, pp. 12\,449--12\,460, 2020.

\bibitem{hsu2021hubert}
W.-N. Hsu, B.~Bolte, Y.-H. Tsai, K.~Lakhotia, R.~Salakhutdinov, and A.~Mohamed, ``Hubert: Self-supervised speech representation learning by masked prediction of hidden units,'' \emph{IEEE/ACM Transactions on Audio, Speech, and Language Processing}, vol.~29, pp. 3451--3460, 2021.

\bibitem{chen2022wavlm}
S.~Chen, C.~Wang, Z.~Chen, Y.~Wu, S.~Liu, Z.~Chen, J.~Li, N.~Kanda, T.~Yoshioka, X.~Xiao \emph{et~al.}, ``Wavlm: Large-scale self-supervised pre-training for full stack speech processing,'' \emph{IEEE Journal of Selected Topics in Signal Processing}, vol.~16, no.~6, pp. 1505--1518, 2022.

\bibitem{fan21_interspeech}
Z.~Fan, M.~Li, S.~Zhou, and B.~Xu, ``{Exploring wav2vec 2.0 on Speaker Verification and Language Identification},'' in \emph{Proc. Interspeech 2021}, 2021, pp. 1509--1513.

\bibitem{yi2021transfer}
C.~Yi, J.~Wang, N.~Cheng, S.~Zhou, and B.~Xu, ``Transfer ability of monolingual wav2vec2. 0 for low-resource speech recognition,'' in \emph{2021 international joint conference on neural networks (IJCNN)}.\hskip 1em plus 0.5em minus 0.4em\relax IEEE, 2021, pp. 1--6.

\bibitem{yang21c_interspeech}
S.~wen Yang, P.-H. Chi, Y.-S. Chuang, C.-I.~J. Lai, K.~Lakhotia, Y.~Y. Lin, A.~T. Liu, J.~Shi, X.~Chang, G.-T. Lin, T.-H. Huang, W.-C. Tseng, K.~tik Lee, D.-R. Liu, Z.~Huang, S.~Dong, S.-W. Li, S.~Watanabe, A.~Mohamed, and H.~yi~Lee, ``{SUPERB: Speech Processing Universal PERformance Benchmark},'' in \emph{Proc. Interspeech 2021}, 2021, pp. 1194--1198.

\bibitem{pasad2021layer}
A.~Pasad, J.-C. Chou, and K.~Livescu, ``Layer-wise analysis of a self-supervised speech representation model,'' in \emph{2021 IEEE Automatic Speech Recognition and Understanding Workshop (ASRU)}.\hskip 1em plus 0.5em minus 0.4em\relax IEEE, 2021, pp. 914--921.

\bibitem{wang22_odyssey}
X.~Wang and J.~Yamagishi, ``{Investigating Self-Supervised Front Ends for Speech Spoofing Countermeasures},'' in \emph{Proc. The Speaker and Language Recognition Workshop (Odyssey 2022)}, 2022, pp. 100--106.

\bibitem{martin2022vicomtech}
J.~M. Mart{\'\i}n-Do{\~n}as and A.~{\'A}lvarez, ``The vicomtech audio deepfake detection system based on wav2vec2 for the 2022 add challenge,'' in \emph{ICASSP 2022-2022 IEEE International Conference on Acoustics, Speech and Signal Processing (ICASSP)}.\hskip 1em plus 0.5em minus 0.4em\relax IEEE, 2022, pp. 9241--9245.

\bibitem{tak22_odyssey}
H.~Tak, M.~Todisco, X.~Wang, J.~weon Jung, J.~Yamagishi, and N.~Evans, ``{Automatic Speaker Verification Spoofing and Deepfake Detection Using Wav2vec 2.0 and Data Augmentation},'' in \emph{Proc. The Speaker and Language Recognition Workshop (Odyssey 2022)}, 2022, pp. 112--119.

\bibitem{huo23b_interspeech}
Z.~Huo, K.~C. Sim, D.~Hwang, T.~Munkhdalai, T.~Sainath, and P.~M. Mengibar, ``{Re-investigating the Efficient Transfer Learning of Speech Foundation Model using Feature Fusion Methods},'' in \emph{Proc. INTERSPEECH 2023}, 2023, pp. 556--560.

\bibitem{vaswani2017attention}
A.~Vaswani, N.~Shazeer, N.~Parmar, J.~Uszkoreit, L.~Jones, A.~N. Gomez, {\L}.~Kaiser, and I.~Polosukhin, ``Attention is all you need,'' \emph{Advances in neural information processing systems}, vol.~30, 2017.

\bibitem{ramachandran2017searching}
P.~Ramachandran, B.~Zoph, and Q.~V. Le, ``Searching for activation functions,'' \emph{arXiv preprint arXiv:1710.05941}, 2017.

\bibitem{nautsch2021asvspoof}
A.~Nautsch, X.~Wang, N.~Evans, T.~H. Kinnunen, V.~Vestman, M.~Todisco, H.~Delgado, M.~Sahidullah, J.~Yamagishi, and K.~A. Lee, ``Asvspoof 2019: spoofing countermeasures for the detection of synthesized, converted and replayed speech,'' \emph{IEEE Transactions on Biometrics, Behavior, and Identity Science}, vol.~3, no.~2, pp. 252--265, 2021.

\bibitem{desplanques20_interspeech}
B.~Desplanques, J.~Thienpondt, and K.~Demuynck, ``{ECAPA-TDNN: Emphasized Channel Attention, Propagation and Aggregation in TDNN Based Speaker Verification},'' in \emph{Proc. Interspeech 2020}, 2020, pp. 3830--3834.

\bibitem{wang2022fully}
C.~Wang, J.~Yi, J.~Tao, H.~Sun, X.~Chen, Z.~Tian, H.~Ma, C.~Fan, and R.~Fu, ``Fully automated end-to-end fake audio detection,'' in \emph{Proceedings of the 1st International Workshop on Deepfake Detection for Audio Multimedia}, 2022, pp. 27--33.

\bibitem{chen21b_asvspoof}
T.~Chen, E.~Khoury, K.~Phatak, and G.~Sivaraman, ``{Pindrop Labs' Submission to the ASVspoof 2021 Challenge},'' in \emph{Proc. 2021 Edition of the Automatic Speaker Verification and Spoofing Countermeasures Challenge}, 2021, pp. 89--93.

\bibitem{chen21_asvspoof}
X.~Chen, Y.~Zhang, G.~Zhu, and Z.~Duan, ``{UR Channel-Robust Synthetic Speech Detection System for ASVspoof 2021},'' in \emph{Proc. 2021 Edition of the Automatic Speaker Verification and Spoofing Countermeasures Challenge}, 2021, pp. 75--82.

\end{thebibliography}

\end{document}